\documentclass[letterpaper, 10 pt, conference]{ieeeconf}  % Comment this line out if you need a4paper

\usepackage{graphicx}
\usepackage{amsmath}
\usepackage{amssymb}
\usepackage{placeins}
\usepackage{subcaption}
\usepackage{caption}
\usepackage{xcolor}
\usepackage{hyperref}
\usepackage{booktabs}
\usepackage{cite}

\usepackage{multirow}

\DeclareMathOperator*{\x}{\mathbf{x}}

\IEEEoverridecommandlockouts                              % This command is only needed if 
\title{\LARGE \bf
Black-box Adversarial Attacks on Network-wide Multi-step Traffic State Prediction Models
}

\author{Bibek Poudel and Weizi Li% <-this % stops a space
\thanks{Bibek Poudel and Weizi Li are with the Department of Computer Science, University of Memphis,
        Memphis, TN 38152, USA 
        {\tt\small \{bpoudel,wli\}@memphis.edu}}%
}

\begin{document}

\maketitle
%\thispagestyle{empty}
%\pagestyle{empty}

%%%%%%%%%%%%%%%%%%%%%%%%%%%%%%%%%%%%%%%%%%%%%%%%%%%%%%%%%%%%%%%%%%%%%%%%%%%%%%%%
\begin{abstract}

Traffic state prediction is necessary for many Intelligent Transportation Systems applications. Recent developments of the topic have focused on network-wide, multi-step prediction, where state of the art performance is achieved via deep learning models, in particular, graph neural network-based models. While the prediction accuracy of deep learning models is high, these models' robustness has raised many safety concerns, given that imperceptible perturbations added to input can substantially degrade the model performance. In this work, we propose an adversarial attack framework by treating the prediction model as a black-box, i.e., assuming no knowledge of the model architecture, training data, and (hyper)parameters. However, we assume that the adversary can oracle the prediction model with any input and obtain corresponding output. Next, the adversary can train a substitute model using input-output pairs and generate adversarial signals based on the substitute model. To test the attack effectiveness, two state of the art, graph neural network-based models (GCGRNN~\cite{Lin2021GCGRNN} and DCRNN~\cite{li2017diffusion}) are examined. As a result, the adversary can degrade the target model's prediction accuracy up to $54\%$. In comparison, two conventional statistical models (linear regression and historical average) are also examined. While these two models do not produce high prediction accuracy, they are either influenced negligibly (less than $3\%$) or are immune to the adversary's attack.

\end{abstract}

%\footnotetext{Second footnote}
\section{Introduction}
\label{sec:intro}

Traffic state prediction is a crucial component of Intelligent Transportation Systems (ITS) and has many applications in traffic control and management. Various statistical and machine learning methods, ranging from traditional techniques~\cite{ahmed1979analysis,anacleto2013multivariate,zhang2013improved} to deep learning models~\cite{lv2014traffic,Lin2019BikeTRB,Lin2021Attention,Lin2021GCGRNN}, have been applied to improve the prediction accuracy. The spatial-temporal resolution of prediction has also evolved from a single timestep on one link~\cite{chandra2009predictions,ma2015long,lin2018quantifying} to multiple timesteps on the entire road network~\cite{cai2016spatiotemporal,Li2017CityEstSparse,Li2018CityEstIter,luo2019multistep,Lin2021GCGRNN}. In this work, we focus on network-wide, multi-step prediction models as they represent the latest advancements regarding the subject.  

While accuracy is an apparent measure of prediction, robustness is another desired feature. A model that can produce high prediction accuracy but acts violently to small perturbations on input has limited use in practice. A better model should be robust under adversarial attacks~\cite{haghighat2020applications}. The demand of robustness is more pronounced during critical conditions such as the pandemic~\cite{Wang2021Mobility,lin2020assessing} and the deep learning era: while deep learning models can deliver state of the art performance, they are known to be fragile to adversarial examples---small crafted perturbations to the input can lead to highly unpredictable behaviors of the model~\cite{szegedy2013intriguing}. As an example, a state of the art deep learning model can mis-classify a stop sign as a 45~mph speed limit sign with minor changes to the sign~\cite{eykholt2018robust}. These types of attacks have serious safety implications for vehicles equipped with such a technology, e.g., autonomous driving.

% Deep Neural Networks (DNNs) have achieved excellent performance in Computer Vision~\cite{krizhevsky2012imagenet,simonyan2014very} and Natural Language Processing~\cite{vaswani2017attention,devlin2018bert}. 
% They are increasingly being adopted in real-world safety-critical domains such as autonomous driving~\cite{PeMS} and medical diagnosis~\cite{hannun2019cardiologist}. Despite the success, state-of-the-art DNNs remain vulnerable to adversarial examples~\cite{szegedy2013intriguing}, where imperceptible perturbations that are added to the input cause a DNN based system to exhibit unexpected behavior. 

% Intelligent Transport Systems (ITS) in recent years has seen large amounts of data being collected by various types of sensors~\cite{nguyen2018deep}. The growth in collected data has driven the use of Deep Learning (DL) in recognition tasks such as detection of traffic signs~\cite{}, vehicle identification~\cite{}, traffic state prediction tasks such as prediction of flow~\cite{}, speed~\cite{li2017diffusion} and other predictive tasks such as travel time calculation~\cite{} and automated ticketing~\cite{}. However, for recognition tasks, an attacker with malicious intent can cause a mis-classification of road traffic signs by applying perturbations that look similar to graffiti art~\cite{eykholt2018robust}. 

% With the increasing adoption of Deep Learning in ITS~\cite{}, the security and privacy of recognition tasks has been widely studied~\cite{}. 

The adversarial attack in ITS has been studied under the context of recognition tasks for traffic signs~\cite{eykholt2018robust} and licence plates~\cite{song2018fooling}, and control and coordination of the vehicle platoon~\cite{dadras2015vehicular}. However, its influence on traffic state prediction has not been receiving equal attention~\cite{haghighat2020applications}. In this paper, we propose an adversarial attack framework to degrade network-wide, multi-step traffic state prediction models by treating the model as a black-box, i.e., we assume no knowledge of the model architecture, training data, and (hyper)parameters. Nevertheless, we assume the adversary can oracle a deployed model (i.e., the target model) using arbitrary input to obtain its output. The input-output pairs are then used to train a substitute model to mimic the target model's behaviors. Next, adversarial signals are generated based on the substitute model via Fast Gradient Sign Method (FGSM)~\cite{goodfellow2014explaining} and Basic Iterative Method (BIM)~\cite{kurakin2016adversarial}, respectively. Using the transferability property~\cite{papernot2016transferability}, the adversary then attacks the target model using the produced adversarial signals with the goal to degrade the target model's prediction performance.  

We test our black-box adversarial attack framework on two deep learning models, namely graph convolutional gated recurrent neural network (GCGRNN)~\cite{Lin2021GCGRNN} and diffusion convolutional recurrent neural Network (DCRNN)~\cite{li2017diffusion}, which correspond to state of the art performance on network-wide, multi-step traffic state prediction. In comparison, we also test two traditional methods, namely linear regression and historical average. Using substitute models trained on the hourly traffic flow data from the Caltrans Performance Measurement System (PeMS)~\cite{PeMS}, we show that the adversary can degrade the performance of GCGRNN up to $26.41\%$ and DCRNN up to $54.07\%$. Compared to deep learning models, traditional models are robust against adversarial attacks with less than $3\%$ or no performance degradation.

% The rest of the paper is organized as follows. Section~\ref{sec:related} introduces related studies regarding traffic state prediction and adversarial attacks in ITS. Section~\ref{sec:prelim} provides the preliminaries of our approach and Section~\ref{sec:method} details the pipeline of our approach. Section~\ref{sec:exp} shows our experiment results, followed by the discuss of the results. The paper concludes by listing some future research directions in Section~\ref{sec:conclusion}. 
\section{Related Work}
\label{sec:related}
In this section, we first introduce relevant deep learning-based studies of network-wide multi-step traffic state prediction. Then, we discuss related studies of adversarial attacks in ITS. 

\subsection{Traffic State Prediction}
Traffic state prediction models need to capture both spatial and temporal dependencies embedded in traffic data for high prediction accuracy~\cite{zhao2019t}. Many deep learning models have been developed for this task. One approach is to use convolutional neural network (CNN). As an example, the traffic state of a city-wide network is converted to a grid map which acts as an ``image'' to be processed by CNN~\cite{ma2017learning}. Since traffic data have varying structures, researchers have adopted a more nature structure---graph---to represent traffic data (e.g., treat traffic sensors as nodes). This has motivated the use of graph convolutional neural network (GCNN)~\cite{cui2019traffic}. One example is diffusion convolutional recurrent neural Network (DCRNN)~\cite{li2017diffusion}, which combines GCNN with recurrent neural network (RNN) to capture the spatial-temporal dependencies. More recently, graph convolutional gated recurrent neural network (GCGRNN) is proposed to integrate data-driven graph filter and gated recurrent neural network for capturing hidden correlations among traffic sensors without requiring a predefined graph representation~\cite{Lin2021GCGRNN}. In this work, we treat GCGRNN and DCRNN, along with two traditional methods (linear regression and historical average), as our target models, and examine their adversarial vulnerability. 

\subsection{Adversarial Attacks in ITS}
Adversarial attacks have been studied extensively for recognition tasks in ITS. Examples include mis-classifying traffic signs at various angles and distances to a vehicle~\cite{eykholt2018robust} and mis-classifying license plates to an optical character recognition system~\cite{song2018fooling}. For control tasks in ITS, there exist studies concerning the coordination of autonomous vehicles. In particular, a single adversarial vehicle is found capable of destabilizing an entire vehicular platoon, causing possible catastrophic accidents~\cite{dadras2015vehicular}. More recently, adversarial attacks are crafted for in-vehicle networks, where a black-box attacker successfully degrades the performance of the deep learning model~\cite{wang2020vulnerability}. While adversarial attacks in ITS have received increasing attention in recent years, to the best of our knowledge, studies that explore the adversarial vulnerability of traffic state prediction models are scarce. In this work, we develop a black-box adversarial attack framework for network-wide, multi-step traffic state prediction models.

\section{Preliminaries}
\label{sec:prelim}
In this section, we briefly introduce the concept of adversarial attack, the attack algorithms used in this work, and the traffic state prediction models considered for attack. 

\subsection{Network-wide Multi-step Traffic State Prediction}
Both deep learning and traditional methods have been used to conduct network-wide, multi-step traffic state prediction. For our attacks, we use two deep learning models and two traditional models as target models. While deep learning models provide state of the art performance, traditional models are easier to interpret and implement. Among traditional models, linear regression is conceptually simple and requires a few parameters to train and historical average is non-parameterized, which requires no training and uses less data and computation for prediction. These models have been widely used in traffic forecasting~\cite{rixey2013station,kamarianakis2012real} and benchmarking~\cite{smith1997traffic,Lin2021GCGRNN}. 

\subsubsection{Diffusion convolutional recurrent neural network (DCRNN)}
DCRNN~\cite{li2017diffusion} requires a predefined spatial graph of sensor networks represented using the adjacency matrix:  
\begin{equation}
A_{ij} = \text{exp}\frac{-dist(s_i,s_j)}{{\sigma}^2},
\label{eq:adj}
\end{equation}
where $dist(s_i, s_j)$ is the spatial distance between the sensors $s_i$ and $s_j$, and $\sigma$ is the standard deviation of the distance.
DCRNN predicts traffic state by capturing a) spatial features of traffic data via random walk on the graph and b) temporal features though an encoder-decoder architecture.

\subsubsection{Graph convolutional gated recurrent neural network (GCGRNN)}
While graph convolution on DCRNN assumes a strong correlation between two sensors that are topologically-close, GCGRNN~\cite{Lin2021GCGRNN} does not make the same assumption. It achieves state of the art performance in traffic state prediction by integrating data-driven graph filter and gated recurrent neural network to automatically learn the adjacency matrix and capture hidden correlations among traffic sensors. In addition, by using gated recurrent unit cells, GCGRNN enjoys better training efficiency.  

\subsubsection{Linear Regression (LR)}
LR estimates the traffic state of the next timestep as a linearly weighted sum of traffic states from historical timesteps. In a network-wide prediction task, each sensor in the network is trained with the ordinary least squares objective (no regularization) and combined over the prediction horizon for estimation.

\subsubsection{Historical Average (HA)}
HA assumes traffic state is periodic: the traffic state of the current period is the average traffic state of previous periods. For example, if we take one week as a period and consider four periods, the predicted traffic state of current Monday is taken as the average state of past four Mondays.

% \subsection{Physical Constraints} 
% % We convert travel times and flows using the road-segment performance function proposed by the Bureau of Public Roads (now Federal Highway Administration) in the U.S.: 

% \begin{equation}
%     t_e=t_{e,min}\left(1+1.5\left(\frac{f_e^4}{c_e}\right)\right), \forall e \in \mathbb{E},
%     \label{eq:a}
% \end{equation}

% \noindent where $c_e$ is the capacity of the road segment $e$, computed as $c_e=1700+10t_{e,min}$ if $t_{e,min} \le 70\text{~mph}$ and $c_e=2400$ otherwise. We set the lower bound of a road segment's travel time $t_e$ as the free-flow travel time, denoted as $t_{e,min}$ and computed using 120\% of the speed limit of the road segment $e$; and the upper bound as the travel time under jam density, denoted as $t_{e,max}$ and computed using the speed 0.5 m/s. 

\subsection{Adversarial Attack}
An adversarial attack on deep neural network $f$ is conducted through crafting an adversarial signal $\x^*$ by imposing minimal perturbation $\delta_{\x}$ to the original signal $\x$.  $\x^*$ can be obtained via solving the following optimization problem~\cite{szegedy2013intriguing}:

\begin{equation}
    \begin{array}{rl}
    {\text{minimize}} & D(\x, \x  + \delta_{\x})\\
    \text{s.t.} & f(\x) \neq f(\x^*),
    \end{array}
    \label{eq:ae}
\end{equation}

\noindent where $D$ is a distance metric such as ${L_0}$, ${L_2}$ or ${L_\infty}$. Multiple methods can be used to solve Eq.~\ref{eq:ae}. Two are examined in this work which are introduced in the following.

\subsubsection{Fast Gradient Sign Method (FGSM)}
FGSM~\cite{goodfellow2014explaining} is designed to produce an adversarial signal in a fast, non-iterative manner. It computes the gradient of the cost function $J$ w.r.t input $\x$ and scales the sign of the gradient by a ${L_{\infty}}$ constraint for generating $\delta_{\x}$:

\begin{equation}
    \mathbf{x}^* = \x + {\varepsilon}{\cdot}sign\left({\nabla}_{x}J\left(\theta,\x,y \right)\right), 
    \label{eq:fgsm}
\end{equation}

\noindent where ${\theta}$ represents the parameters of $f$; $y$ is the target; and ${\varepsilon}$ is the ${L_{\infty}}$ constraint parameter controlling the attack magnitude. Increasing the value of ${\varepsilon}$ will increase the attack effectiveness but also make the adversarial example more distinguishable (i.e., having large $\delta_{\x}$).

\subsubsection{Basic Iterative Method (BIM)}
BIM~\cite{kurakin2016adversarial} is a refined, iterative method of FGSM that generates an adversarial signal. At each iteration, BIM will take a small step $\alpha$ and clip the result using ${\varepsilon}$:

\begin{equation}
\mathbf{x}^*_{n+1} = clip_{\x,{\varepsilon}}
\left(\mathbf{x}_n^*+\alpha\cdot sign\left(\nabla_{\x} J\left(\theta,{\mathbf{x}^*_n},y\right)\right)\right),
\label{eq:bim}
\end{equation}

\noindent where $\theta$ denotes the parameters of $f$. Compared to FGSM, BIM can produce a less distinguishable adversarial signal compared to the original signal $\x$ at a cost of more computation. 
\section{Methodology}
\label{sec:method}

Our attack takes the form of a black-box threat model, where an adversary is assumed to possess no knowledge about the target model's architecture, training data, and (hyper)parameters. However, the adversary can oracle the target model by feeding any input data to it and obtain corresponding output. The goal of the adversary is then to perturb the input data so that the target model's prediction accuracy will be degraded. The perturbed input is the adversarial signal and can be obtained by solving the optimization program in Eq.~\ref{eq:ae} based on a substitute model that mimics the target model's behavior.   

\begin{figure}[ht]
		\centering
		\includegraphics[width=.9\linewidth]{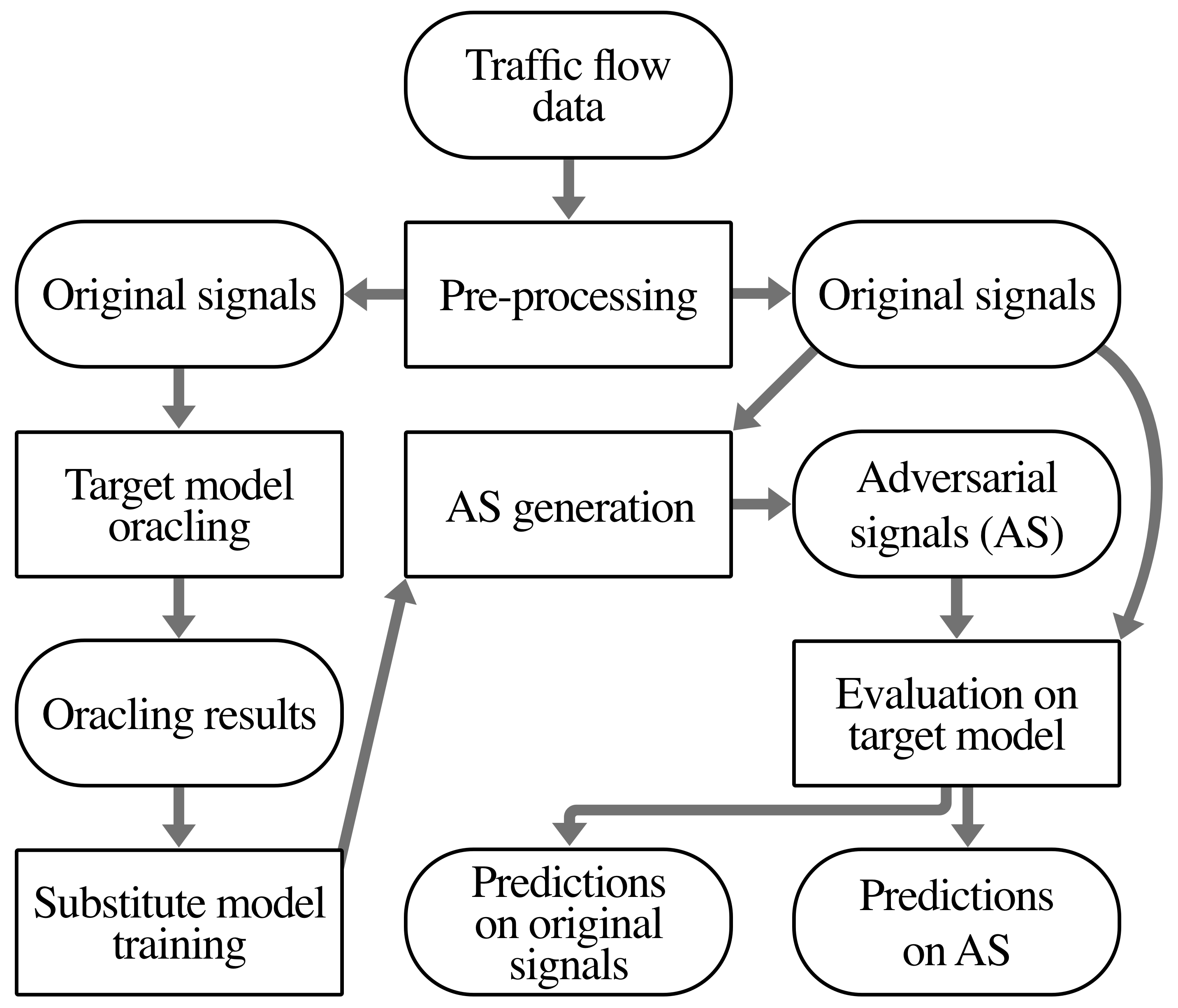}
		\caption{Systematic diagram of our framework. Network-wide traffic flow data is pre-processed and split into two parts, one is used by the adversary to oracle the target model to obtain input-output pairs for training the substitute model; the other is used by the adversary to generate adversarial signals based on the trained substitute model. The attack effectiveness is conducted on the target model using original signals and their corresponding adversarial signals.}
		\label{fig:systematic}
\end{figure}

The systematic diagram of our framework is shown in Fig.~\ref{fig:systematic}. The network-wide traffic flow data is pre-processed and split into input data for oracling the target model and ground-truth data for later evaluation of the attack effectiveness. The adversary feeds the input data into the target model and obtains corresponding output predictions. These input-output pairs are then used to train and tune a substitute model such that the substitute model produces similar prediction performance as the target model. Once this is done, adversarial signals are generated using the substitute model to attack the target model. In the following, we detail these steps.       

% flow data is downloaded from PeMS, 150 sensors across a road network are selected and pre-processed by applying 12 time step window approach. The obtained inputs are fed to the target model to obtain predictions. The network-wide input and prediction pairs are used as training data for a substitute CNN model. 

% \begin{figure}
% 		\centering
% 		\includegraphics[width=.75\linewidth]{figures/map_3.png}
% 		\caption{The locations of the 150 sensors in Los Angeles used in our study. Most of the sensors are located on highways.}
% 		\label{fig:sensors}
% \end{figure}

\begin{table*}[ht]
    \centering
    \caption{LEFT: Average L2 distance between the 2~616 original signals and their adversarial signals as well as average L2 distance between the target model's predictions on 2~616 original signals and their corresponding adversarial signals. All values are normalized by $10^8$. RIGHT: Performance evaluation of target models before and after the attack. Deep Learning models suffer severe degradation in terms of their prediction performance, compared to traditional models. DCRNN has the highest performance degradation among all models.} 
    \scalebox{1}{
    \begin{tabular}{ccccc|ccccc}
        \toprule
        & \multicolumn{2}{c}{Change on original signal (L2)}  & \multicolumn{2}{c}{Change on prediction (L2)} & \multicolumn{1}{c}{Pre-attack RMSE} & \multicolumn{2}{c}{Post-attack RMSE}  & \multicolumn{2}{c}{RMSE Degradation(\%)} \\ 
        \cmidrule(l){2-3} \cmidrule(l){4-5} \cmidrule(l){6-6} \cmidrule(l){7-8} \cmidrule(l){9-10}
        Model & FGSM & BIM & FGSM & BIM & N/A &  FGSM & BIM & FGSM & BIM \\
        \midrule %Table head should appear above the table
        GCGRNN & 3.35 & 1.67 & 3.27& 1.80 & 529.88 &  669.87 & 607.14 & 26.41\% & 14.58\% \\
        \midrule
        DCRNN & 3.35 & 1.45 & 17.5& 10.1 & 770.04 &  1186.41 & 1013.66 & 54.07\% & 31.63\% \\
        \midrule
        LR & 3.35 & 1.45 & 3.16 & 1.38 & 1870.43 & 1916.83 & 1891.21 & 2.48\% & 1.11\% \\
        \midrule
        HA & 132 & 129 & 0 & 0 & 935.46 & 935.46 & 935.46 & 0 & 0 \\ 
        \bottomrule
    \end{tabular}}
    
    \label{tab:master_table}
\end{table*}

%used in our study. Most sensors are located on highways Sensorswith the average hourly traffic volume in the top 25 percentilesare red; sensors with the average hourly traffic volume in thebottom 25 percentiles are black. The rest are blue. The sensorsare  mainly  located  along  highways.  Sensors  located  at  thewest  side  of  the  network  show  higher  traffic  volumes  thanthe sensors located at the east side of the network.%

\subsubsection{Data pre-processing}
The traffic flow data is obtained from PeMS~\cite{PeMS}, which consists of hourly data from Jan 1, 2018 to Jun 30, 2019 of 150 sensors in Los Angeles, California. To prepare this network-wide data for multi-step prediction, 12 timesteps of data in the past are treated as inputs and next 12 timesteps of data are treated as outputs (i.e., ground truth). In total, we obtain 13~081 network-wide, 12-step traffic flow input-output data.

\subsubsection{Substitute model training}
In order to train the substitute model, 9~157 (sampled from 13~081) input data is used to oracle the target model to obtain the corresponding predictions (which are different from the ground-truth output data for the 9~157 input data). These input-output pairs form our training data for the substitute model, which are z-score normalized using:

\begin{equation} 
    \x = (\x - {\mu})/ {\sigma},
    \label{eq:norm}
\end{equation}

\noindent where ${\mu}$ and ${\sigma}$ are the mean and standard deviation of the 9~157 input data, respectively.

The substitute model is chosen to be the 50-layered residual network (ResNet-50)~\cite{he2016deep} as it achieves state of the art performance in time series classification on various UCR datasets~\cite{wang2017time}. To make the model more suitable for our purpose, we alter the model's architecture by removing the pooling operation from initial layers to better preserve temporal correlations embedded in traffic data, and change the input and output layers to match the prediction task of 150 sensors across 12 timesteps. For each target model examined (four in total), a substitute model with randomly initialized parameters and the same set of hyperparameters is trained, using the 9~157 input-output pairs.

\subsubsection{Adversarial attack}
In total, 2~616 adversarial signals are generated using both Fast Gradient Sign Method (FGSM)~\cite{goodfellow2014explaining} and Basic Iterative Method (BIM)~\cite{kurakin2016adversarial} by constraining the optimization program given in Eq.~\ref{eq:ae} with the maximum and minimum values of normalized inputs obtained from Eq.~\ref{eq:norm}. This ensures that the adversarial signals are physically plausible, i.e., no negative traffic flows should occur. The adversarial signals generated on the substitute model are fed into the target model to obtain the ``post-attack predictions'' as the attack results. To measure the attack effectiveness, the inputs that are used to generate these adversarial signals are also fed into the target model to obtain ``pre-attack predictions.'' Root mean squared error (RMSE) is used as an evaluation metric to measure the model performance:

\begin{equation} 
    \textrm{RMSE} = \sqrt{\frac{1}{S\times F \times N} {\sum_{{i}}^S} {\sum_{{d}}^F}{\sum_{{n}}^N}{(y^i_{dn} - \hat{y}^i_{dn})}^2},
    \label{eq:8}
\end{equation}

\noindent where $S$ = 2~616 is the number of adversarial signals; $F$ = 12 is the number of timesteps; $N$ = 150 is the number of sensors; $y_{dn}$ is the ground-truth data from PeMS and $\hat{y}_{dn}$ is either the ``post-attack prediction'' or ``pre-attack predcition''. Denoting $\text{RMSE}_{pre}$ the difference between ``pre-attack predictions'' and the ground-truth, and $\text{RMSE}_{post}$ the difference between ``post-attack predictions'' and the ground-truth, the performance degradation is calculated as

\begin{equation} 
    \textrm{RMSE Degradation} = \frac{\textrm{RMSE}_{post} - \textrm{RMSE}_{pre}}{\textrm{RMSE}_{pre}}. 
    \label{eq:10}
\end{equation} 

\begin{figure*}[pt]
		\centering
		\includegraphics[width=\textwidth]{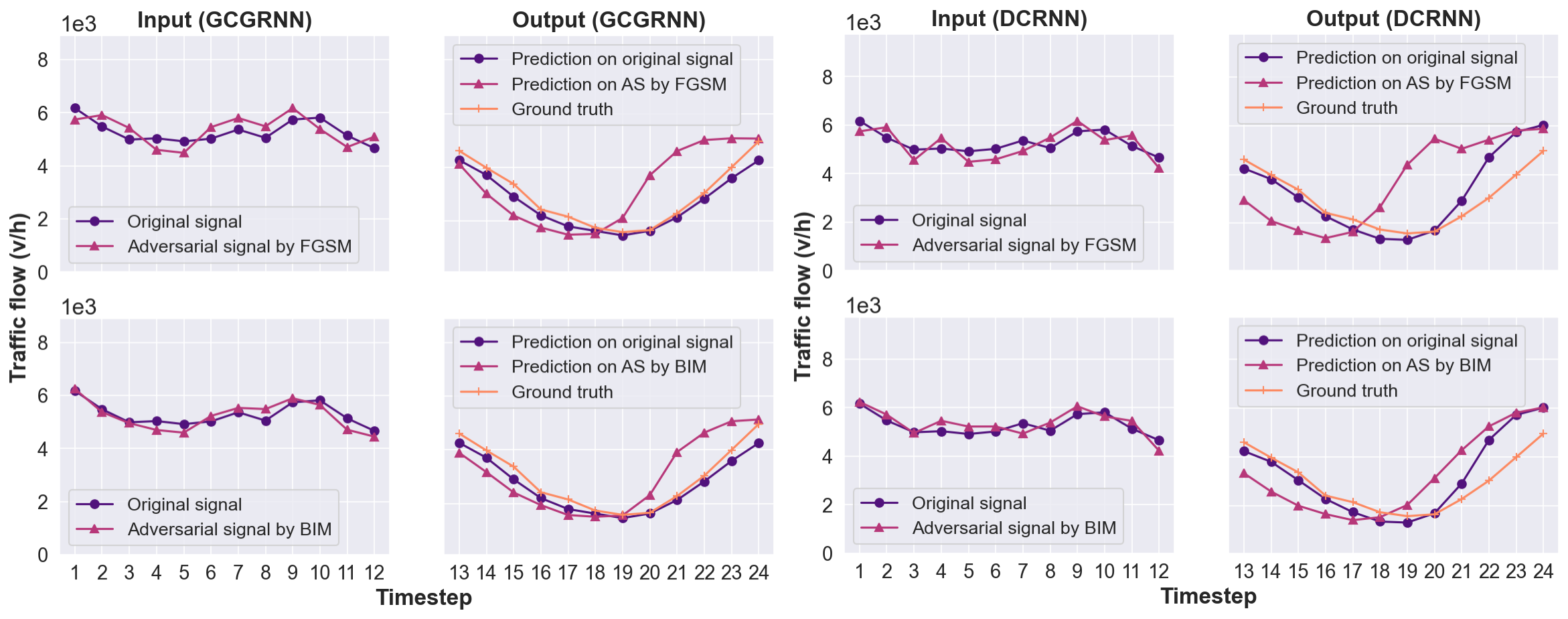}
		\caption{Examples of original signals and their corresponding adversarial signals' impact on model prediction results. For both GCGRNN and DCRNN, we can see that the adversarial signals can cause large deviation between the prediction results and the ground truth than the original signals.} 
		\label{fig:signals}
\end{figure*}

% (a) A 12-step original signal in our dataset and its corresponding adversarial signals from FGSM and BIM are respectively shown as Input (b) Ground truth and model predictions (GRGRNN, DCRNN) for the next 12 timesteps on both the original signal and adversarial signals are shown as GCGRNN/ DCRNN predictions. Under no attack, GCGRNN has the lowest RMSE among all target models i.e., ground truth and predictions on original signal are close however the predictions on adversarial signals maximize RMSE by moving predictions further away from ground truth. The difference between ground truth and prediction on AE is more pronounced in DCRNN prediction.

\begin{figure*}[pt]
		\centering
		\includegraphics[width=\textwidth]{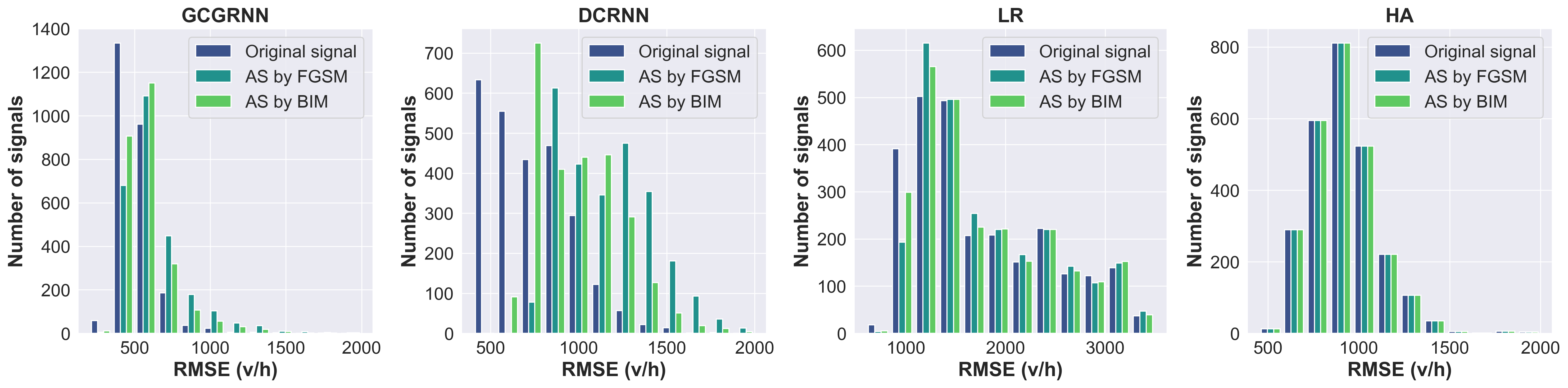}
		\caption{The RMSE distributions on both original signals and their adversarial signals (AS) of four target models ($2~616$ in total). Between GCGRNN and DCRNN, DCRNN shows larger distribution shift in RMSE values, which demonstrates the comparative robustness of GCGRNN against adversarial attacks. For LR, the shift is marginal; for HA, no shift is observed. These results demonstrate that traditional models can be more robust or even immune against adversarial attacks. } 
		\label{fig:distributions}
\end{figure*}

\section{Experiments}
\label{sec:exp}

In this section, we first describe the experiment set-up and then present and discuss the experiment results. All experiments are conducted using an Intel(R) Core(TM) i7-10700 CPU, a Nvidia RTX 2080 SUPER GPU, and 32G RAM. PyTorch~\cite{paszke2017automatic} is used to implement the substitute model and Advertorch~\cite{ding2019advertorch} is used to generate adversarial signals. 

The training data obtained by oracling the four pre-trained target models is respectively used to train four substitute models under the objective to minimize RMSE. The values of the hyperparameters are as follows: learning rate $= 0.005$, optimizer $=$ Adam~\cite{kingma2014adam}, batch size $= 24$, number of epochs $= 50$. To generate adversarial signals, RMSE is used as the cost function for both FGSM (with $\varepsilon = 0.2$) and BIM (with $\alpha = 0.05,~\varepsilon = 0.2$, number of iterations $= 10$). The upper and lower limit constraints on the optimization program in Eq.~\ref{eq:ae} are set to $3.49$ and $-1.89$, respectively.

Table~\ref{tab:master_table} LEFT shows the results of using L2 distance as the metric to measure 1) the difference between the original signal and its corresponding adversarial signal and 2) the difference between the target model's prediction on the original signal and the corresponding adversarial signal. Except for HA, the changes produced on the original signal by FGSM ($3.35\times 10^8$) are almost doubled than that of BIM ($1.67\times 10^8$, $1.45\times 10^8$), indicating that the adversarial signal from BIM are in general closer to the original signals than those from FGSM. For HA, the changes produced by FGSM ($1.32\times 10^{10}$) and BIM ($1.29\times 10^{10}$) to the original signals are much higher. 

This may due to the inability of the substitute model to mimic HA's behaviour: HA makes a periodic assumption on traffic state prediction i.e., to make a prediction on next 12 timesteps from an input of 12 historical timesteps, it does not use the input flow values, rather it looks at the hours of day, and day of the week (assuming period = week) of the input and averages the flow values over the same hours of day on the same day of the week from previous weeks. The substitute model makes predictions based on the learned features from the flow values of 12 historical timesteps, which does not contain periodic information. In other words, there is no way to tell if two network-wide multi-step inputs belong to the same period (e.g., the same hours of the same day of week). As a result, the substitute model fails to learn HA's behavior, causing FGSM and BIM to produce large changes to the input signal while maximizing RMSE.

Regarding the changes on predictions, for GCGRNN and LR, changes under FGSM ($3.27\times 10^8$, $3.16\times 10^8$) are almost doubled than that of the changes under BIM ($1.80\times 10^8$, $1.38\times 10^8$). These changes are similar to the changes produced on the original signals to craft adversarial signals using FGSM and BIM respectively. In contrast, no changes are produced on the predictions for HA, this is again explained by the periodic assumption of HA since HA does not consider the flow values in the current input rather it considers the historical periods that it belongs. Hence, changes produced on current input cannot influence the prediction.

DCRNN suffers the largest changes at $17.5\times 10^8$ under FGSM and $10.1\times 10^8$ under BIM, indicating that the substitute model highly resembles DCRNN. Fig.~\ref{fig:signals} shows the predictions of GCGRNN and DCRNN on both original signals and their corresponding adversarial signals from either FGSM or BIM. As shown, the adversarial signals can cause much larger deviation between the model prediction and the ground truth than that of the original signals.  

Table~\ref{tab:master_table} RIGHT gives information about the target model predictions before the attack, after the attack, and RMSE degradation. Under no attack, GCGRNN has the smallest RMSE of $529.88$, which has increased to $699.87~(26.41\%~\text{RMSE degradation})$ using FGSM and $607.14~(14.58\%~\text{RMSE degradation})$ using BIM under adversarial attacks, respectively. The highest performance drop is for DCRNN where the RMSE of $770.4$ under no attack has increased to $1186.41~(54.07\%~\text{RMSE degradation})$ using FGSM and $1013.66~(31.63\%~\text{RMSE degradation})$ using BIM under adversarial attacks, respectively. The RMSE distributions on both original signals and their adversarial signals are shown in Fig.~\ref{fig:distributions}. DCRNN corresponds to more dramatic distribution shift to higher RMSE values than GCGRNN, which demonstrates the comparative robustness of GCGRNN against adversarial attacks than DCRNN.

For traditional models' performance shown in Table~\ref{tab:master_table} RIGHT, LR operates at high RMSE under no attack ($1870.43$) because of the linearity assumption (linearly combines historical timesteps for prediction). The RMSE has increased to $1916.83~(2.48\%~\text{RMSE degradation})$ using FGSM and $1891.21~(1.11\%~\text{RMSE degradation})$ using BIM under adversarial attacks, respectively. The ``RMSE degradation'' values for LR are relatively small compared to that for GCGRNN and DCRNN. From Table~\ref{tab:master_table} LEFT, we can see that although the ``change on prediction'' under both FGSM and BIM for LR are similar to ``change on prediction'' for GCGRNN, the difference between ``Pre-attack RMSE'' and ``Post-attack RMSE'' is higher for GCGRNN than LR. This indicates that for LR, although the predictions on adversarial signals are far from predictions on original signals, their distances to the ground truth remain similar, i.e., the LR's prediction already has high RMSE due to the model limitation in capturing temporal correlations embedded in historical traffic flow. In case of HA, the performance remains invariant to any attack, i.e., RMSE remains the same for original signals and adversarial signals due to the periodicity assumption explained before. 

The lower prediction performance of both traditional models under no attack, due to their inability to accurately capture spatial and temporal correlations embedded in traffic data (owing to their simplistic assumptions of either linearity (LR) or periodicity (HA)), may be a contributor to their robustness. As can seen from Fig.~\ref{fig:distributions}, there exists marginal distribution shift to higher RMSE values for LR and no distribution shift for HA.

\section{Conclusion and Future Work}
\label{sec:conclusion}
% - adversarial examples for a sequence to sequence prediction problem in a regression setting (no one has done it, people have done AEs in NLP where there are sequence to sequence scenarios)
% - robustness research

% systematic diagram again
% more detailed than abstract
% add yourself (above)

In this work, we propose a black-box adversarial attack framework where an adversary can attack traffic state prediction models using publicly available dataset to oracle a target model, train a substitute model on prediction results, and attack the target model using the produced adversarial signals based on the substitute model. We have analyzed two deep learning models and two traditional models and find that traditional models are more robust to the proposed adversarial attack. Among the deep learning models, we find that GCGRNN is comparatively more robust against adversarial attacks than DCRNN.  

There are many future research directions of this work. First of all, the input data for oracling the target model is complete and accurate. It would be interesting to study the influence of compressing data~\cite{lin2019efficient} or even missing data~\cite{Li2017CityFlowRecon} on the attack. Secondly, our work could be extended to adversarial attacks on other traffic measurements such as speed and travel time. Another interesting topic is to study the impact of adversarial attack on existing techniques to navigate and coordinate autonomous vehicles~\cite{wu2018stabilizing,Li2019ADAPS}, which heavily rely on accurate prediction of traffic states to operate. This line of research can benefit from the use of virtual traffic platforms~\cite{Wilkie2015Virtual,Chao2020Survey} for experimenting, since the real-world deployment of the algorithms has safety implications. 

Last but not least, deep learning models (e.g., GCGRNN and DCRNN) are data-driven, which in general do not consider domain knowledge such as short-term seasonality (e.g., weekdays vs. weekend), long-term seasonality (e.g., Spring vs. Summer), and special events (e.g., holidays). Given that the real-world traffic is associated with such information, it would be worthwhile to explore their inclusion as a part of the input.

%\section*{ACKNOWLEDGMENT}
%TBD

% \begin{thebibliography}{99}
% \clearpage
% \FloatBarrier
\bibliographystyle{IEEEtran}
\bibliography{ref}

\end{document}